\documentclass[letterpaper,10pt]{article} 
\usepackage[utf8]{inputenc}

\usepackage{array}
\usepackage{algorithmic}
\usepackage{algorithm}
\usepackage{makecell}
\usepackage{float}
\usepackage{booktabs}

\usepackage{longtable}

\usepackage{amsmath,amssymb}
\usepackage[colorlinks=true,bookmarks=false,citecolor=blue,urlcolor=blue]{hyperref} 

\usepackage{bbding}
\usepackage{adjustbox}
\usepackage{tabularx}

\usepackage{graphicx}

\DeclareUnicodeCharacter{0301}{\'{e}}
\usepackage{multirow}
\usepackage{titlesec}

\usepackage{setspace}

\providecommand{\keywords}
{
\textbf{\text{Keywords:}}
}

\usepackage[T1]{fontenc}

\usepackage{authblk}

\title{\vspace{-1cm}\textbf{CPP-DIP: Multi-objective Coverage Path Planning for MAVs in Dispersed and Irregular Plantations}}
\author[1]{\textbf{Weijie Kuang}}
\author[1]{\textbf{Hann Woei Ho} \thanks{Corresponding author: aehannwoei@usm.my}}
\author[1]{\textbf{Ye Zhou}}

\affil[1]{School of Aerospace Engineering, Engineering Campus, Universiti Sains Malaysia, 14300 Nibong Tebal, Pulau Pinang, Malaysia}

\date{}

\usepackage{geometry}

\begin{document}

\maketitle

\begin{singlespacing}

\begin{abstract}
Coverage Path Planning (CPP) is vital in precision agriculture to improve efficiency and resource utilization. In irregular and dispersed plantations, traditional grid-based CPP often causes redundant coverage over non-vegetated areas, leading to waste and pollution. To overcome these limitations, we propose CPP-DIP, a multi-objective CPP framework designed for Micro Air Vehicles (MAVs). The framework transforms the CPP task into a Traveling Salesman Problem (TSP) and optimizes flight paths by minimizing travel distance, turning angles, and intersection counts. Unlike conventional approaches, our method does not rely on GPS-based environmental modeling. Instead, it uses aerial imagery and a Histogram of Oriented Gradients (HOG)-based approach to detect trees and extract image coordinates. A density-aware waypoint strategy is applied: Kernel Density Estimation (KDE) is used to reduce redundant waypoints in dense regions, while a greedy algorithm ensures complete coverage in sparse areas.
To verify the generality of the framework, we solve the resulting TSP using three different methods: Greedy Heuristic Insertion (GHI), Ant Colony Optimization (ACO), and Monte Carlo Reinforcement Learning (MCRL). Then an object-based optimization is applied to further refine the resulting path. Additionally, CPP-DIP integrates ForaNav, our insect-inspired navigation method, for accurate tree localization and tracking. The experimental results show that MCRL offers a balanced solution, reducing the travel distance by 16.9 \% compared to ACO while maintaining a similar performance to GHI. It also improves path smoothness by reducing turning angles by 28.3 \% and 59.9 \% relative to ACO and GHI, respectively, and effectively eliminates intersections. These results confirm the robustness and effectiveness of CPP-DIP in different TSP solvers.

\keywords{Micro Air Vehicles; Coverage Path Plan; Traveling Salesman Problem; Object Detection; Autonomous Navigation.}
\end{abstract}

\section{Introduction}
Coverage Path Planning (CPP) plays is key in precision agriculture operations, ranging from fruit harvesting, pesticide spraying, and fertilizing to growth monitoring \cite{hoffmann2024optimal}. It involves planning efficient routes for mobile agents to maximize coverage while optimizing resource use \cite{wu2025full, dogru2022eco}. Although CPP has been well-established in densely planted and organized agricultural regions, challenges still arise in dispersed and irregularly planted regions. 

There are two main challenges, particularly when using resource-limited agents, such as Micro Air Vehicles (MAVs), where energy efficiency is crucial for implementation \cite{lee2023air, song2024energy}. MAVs are lightweight, highly maneuverable, and capable of operating at low altitudes, making them well-suited for precision agriculture tasks such as crop monitoring and targeted spraying \cite{ijaz2024robust, divazi2025experimental}. The first issue is that the generated paths may cover non-vegetated areas, resulting in redundant paths and wasted chemical usage. Usually, target-based path planning solves this problem by converting the traditional grid-based CPP into a Traveling Salesman Problem (TSP) \cite{wei2024precise, tian2023design, wang2023coverage}. This transformation is based on the coordinate information of plants. TSP is an optimization problem that aims to find the optimal route that visits each point exactly once and returns to the beginning \cite{zhou2023bi, wang2024collaborative}.

Specifically, target-based path planning shifts the focus from generating paths for the entire area to planning paths based on target locations \cite{lu2025target}. This shift makes environmental modeling essential for obtaining the spatial coordinates of each crop. The second challenge is that environmental modeling typically depends heavily on GPS. This requires significant time and extra equipment, making it costly and unsuitable for large-scale planting areas.

In traditional TSP, the total travel distance is the primary metric for evaluating path quality \cite{jia2018cooperative, xu2025discrete}. However, when MAVs are used to perform CPP-based tasks, additional factors, such as turning angles and intersecting paths, become equally critical \cite{rezaee2024comprehensive}. Large turning angles can significantly impact flight duration, while intersecting paths may result in overlapping applications, leading to overspraying and potential environmental pollution.

Therefore, we propose a multi-objective CPP framework, DIP-CPP, for MAVs operating in dispersed and irregular plantations. It incorporates path angles and intersection counts as additional optimization objectives. This approach requires only an aerial image of the planting area. It avoids redundant paths, reduces the number of waypoints, and eliminates the need for extensive pre-planning environmental modeling. This study selects oil palm plantations as the experimental site due to their significant economic value. Their high-density planting environment often suffers from poor management, which will lead to tree dispersion \cite{khuzaimah2022application}. The proposed framework transforms traditional grid-based CPP into TSP and applies it to three common algorithms for comparative performance analysis. The framework was further validated experimentally using an MAV as the mobile agent.

The \textit{main contributions} of this paper are:
\begin{itemize}
\item a multi-object coverage path planning framework integrating travel distance, turning angle, and intersection count for the efficient path,
\item a density-aware coverage method utilizing Kernel Density Estimation (KDE) to handle high-density tree regions, reducing unnecessary waypoint assignment,
\item an object-optimized path replanning strategy that selects the best waypoints within a feasible region and
\item integration with our previous work, ForaNav \cite{kuang2025foranav}, for real-time tree detection, enabling precise MAV positioning above trees during waypoint navigation.
\end{itemize}

The remainder of this paper is organized as follows: Section \ref{sec:relatedworks} discusses the recent relevant literature. In Section \ref{sec:method}, the proposed method is detailed. Section \ref{sec:results} outlines the flight experiments and results discussion, and finally, the conclusion is drawn in Section \ref{sec:conclusion}.

\section{Related works}
\label{sec:relatedworks}
In precision agriculture with the need for advanced automation, CPP is critical in optimizing resource use, minimizing soil compaction, and enhancing crop yields \cite{thakar2022area, krestenitis2024overcome}. In recent years, CPP has garnered significant interest from researchers \cite{li2025optimal}. For example, inspired by the predator-prey relation, an adaptive CPP approach was introduced to respond efficiently to dynamic environments \cite{hassan2019ppcpp}. The prey continuously maximizes its distance from the predator while covering the target area and avoiding obstacles at minimal cost. To reduce turn frequency in the path, a concave point elimination-based convex decomposition algorithm was applied in CPP \cite{jia2024coverage}. It also minimizes computational costs by merging subconvex regions and using weighted minimum traversal for path generation. Moreover, another study explored multi-objective optimization in CPP to enhance overall performance \cite{fu2023full}. It considered constraints like coverage rate, overlap rate, trajectory length, and cumulative turning angle. To optimize the solution space, a coordinate system transformation method was introduced. However, these studies did not consider path intersections. This is crucial in agricultural applications, such as pesticide spraying, where intersections can lead to contamination and resource waste.

To effectively reduce costs and improve adaptability in large-scale areas, there are also some studies focused on multi-agent systems \cite{zhu2024multi}. The CPP for autonomous heterogeneous vehicles in a bounded number of regions was investigated \cite{chen2021clustering}. It proposed both an exact mixed integer linear programming formulation for optimal flight paths and a clustering-based algorithm for efficient point-to-point path generation. Another research developed a multi-agent system for CPP across multiple separate areas, featuring the ability to re-plan missions in response to unexpected events \cite{luna2024multi}. For tasks in dynamic environments, another study proposed an Artificial Potential Field-based (APF) method to improve multi-robot collaboration. Instead of directly generating paths from the field gradient, it utilizes APF to define coverage policies \cite{wang2024apf}. Despite these studies showing good performance, they are grid-based CPPs with full area coverage. As a result, when dealing with dispersed and irregular planting regions, covering non-vegetated areas becomes unavoidable.

To minimize irrelevant coverage, prevent pollution, reduce flight time, and further improve coverage efficiency, some studies focused on target-based CPP approaches \cite{wei2024precise, tian2023design, wang2023coverage}. They typically begin by detecting objects to obtain their coordinates which serve as waypoints. Next, the CPP is converted into a TSP to traverse all waypoints based on a specific goal, such as minimizing the distance.
For example, a study applied Ant Colony Optimization (ACO) to optimize the shortest path. This enabled the agent to navigate each fruit tree within the experimental area \cite{wei2024precise}. This approach achieves a 2.04 \% reduction in distance compared to the grid-based CPP method. To combine corner and distance costs in path planning and speed up the iteration of ACO, another research proposed the Multi-Source ACO (MS-ACO) \cite{tian2023design}. Compared to ACO, MS-ACO significantly reduces the total path length and turning angle. However, these studies were conducted in well-managed plantations, leaving their performance in dispersed regions uncertain. 

Reinforcement learning has also been applied for CPP \cite{boulares2024uav}. For example, a deep reinforcement learning-based CPP method was developed for a picking robot to harvest dispersed kiwifruit \cite{wang2023coverage}. The results showed the path length of this method was shorter 31.56  \% than that of the grid-based CPP method. This study was conducted with a ground picking robot and did not focus on its endurance. Therefore, path turning angles and energy consumption were not important. Meanwhile, these methods also did not consider intersections and spent significant time to acquiring target locations.

To solve the CPP problem using energy-limited mobile agents in dispersed and irregular plantations, we develop a CPP framework incorporating path length, turning angles, and the number of path intersections as optimization objectives. The framework first utilizes a single aerial image of the plantation to obtain the image coordinate of trees. These coordinates are derived from the tree detection results using an HOG-based detection method. Based on image coordinates, multiple cover circles with a radius equal to the field of view of the onboard camera are used, where the centers of these circles serve as waypoints to cover all trees. The Greedy Algorithm then selects optimal positions for these cover circles to minimize the number of waypoints. Hereby, the CPP problem is thus transformed into the TSP. Three common algorithms for solving TSP are subsequently applied to plan the coverage path. Finally, an object-based optimization is employed to refine the planned path further.

\section{Methodology}
\label{sec:method}
\begin{figure}[H]
    \centering
    \includegraphics[width=4.5in]{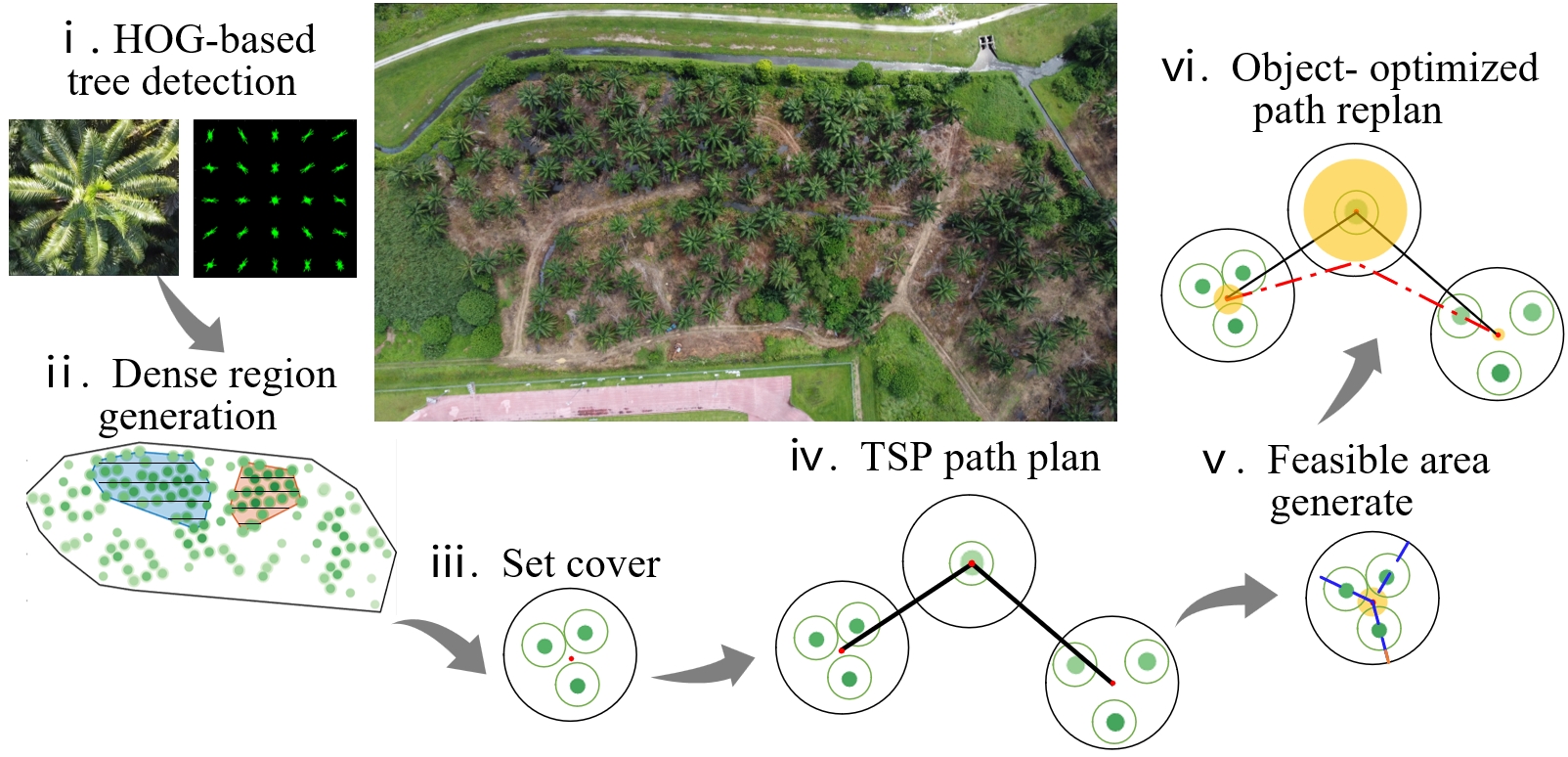}
    \caption{The framework of CPP-DIP. Tree image coordinates are derived from detection results obtained through a HOG-based method. Next, high-density regions are segmented using a KDE approach, while cover circles are produced in low-density areas using a greedy algorithm to generate waypoint coordinates. A path is then planned through these waypoints by solving the TSP. Finally, this path is further refined through object-based optimization.}
    \label{Flowchart}
\end{figure}

As shown in Fig. \ref{Flowchart}, this section presents our CPP framework composed of four main parts: Section \ref{sec:detection}  \textit{Oil palm tree detection}, Section \ref{sec:Waypoint} \textit{Waypoint and dense region generation}, Section \ref{sec:PathPlan} \textit{Target-based path planning}, and Section \ref{sec:RePlan} \textit{object-optimized path replanning}. Tree detection derives tree image coordinates from an aerial image, providing the necessary spatial information for waypoint generation. Based on this, a density-aware waypoint generation strategy is applied, utilizing Kernel Density Estimation (KDE) and Density-Based Spatial Clustering of Applications with Noise (DBSCAN) clustering to reduce waypoints in high-density regions while ensuring sufficient coverage in low-density areas. Finally, path planning and object-optimized path replanning refine the coverage path to balance distance, turning angles, and intersection counts for efficient MAV navigation.

\subsection{Oil palm tree detection}
\label{sec:detection}
We adopted the detection method from the detection module in our previous research, ForaNav \cite{kuang2025foranav}. This method utilizes the HOG feature descriptor to extract star-shaped features of oil palm trees and classifies them using an SVM model. Specifically, ForaNav employs a hierarchical feature extraction method to enhance oil palm tree detection. The bottom layer captures fine needle-like leaflets using high-resolution images, while the top layer extracts star-shaped features from down-sampled images. The variance of the bottom-layer HOG features helps distinguish oil palms from similar species. An SVM classifier is then trained to recognize star-shaped structures. In addition, hue and saturation histograms are analyzed using the Pearson correlation coefficient to differentiate oil palms from background objects with similar edge patterns. ForaNav has been proven in previous research to be effective for real-time detection in cluttered environments.

Furthermore, ForaNav is applied beyond detection in waypoint-based navigation. By integrating real-time tree tracking, ForaNav enables MAVs to adjust their flight paths dynamically, enabling precise positioning above target trees. This adaptive navigation approach reduces deviations caused by environmental disturbances or localization errors. This capability is particularly beneficial for precision agriculture tasks. Accurate tree localization is crucial for effective intervention, such as targeted spraying or automated fruit ripeness detection.

\subsection{Waypoint and dense region generation}
\label{sec:Waypoint}
Based on the obtained tree image coordinates derived from detection results, all trees can be covered using multiple cover circles. This approach ensures that all trees within the circle can be seen once the MAV flies to the center of the cover circle. In this study, oil palm trees are represented as small circles with a radius $r$, while the field of view is represented by larger circles with a radius $R$. 
The Greedy Algorithm is used for this geometric coverage problem due to its simplicity, as it selects the best option at each step.

\begin{figure}[!t]
    \centering
    \includegraphics[width=4.5in]{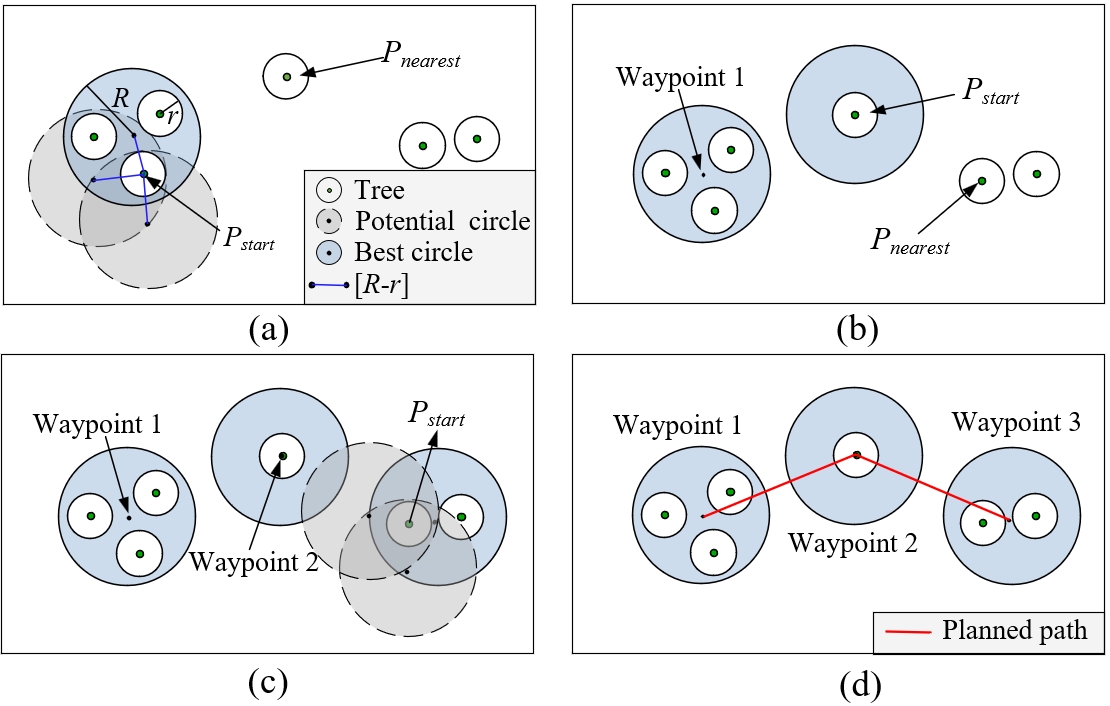}
    \caption{Coverage-based waypoint generation. (a) Search in potential region; (b) First waypoint generation; (c) Sequential waypoint generation; (d) Path plan based on waypoints.}
    \label{SetCover}
\end{figure}

Algorithm \ref{alg1} takes the image coordinates of trees as input and aims to generate large circles that cover all trees, with the centers of these circles serving as waypoints. Fig. \ref{SetCover} illustrates how the algorithm iteratively determines the positions for these large circles to cover a set of small circles, aiming to minimize the number of large circles \cite{wang2023coverage}. Initially, the algorithm calculates the Euclidean distance between each tree and randomly selects a tree to serve as the starting point $\textbf{\emph{P}}_{start}$. A search is then conducted among all potential positions around $\textbf{\emph{P}}_{start}$. The goal is to determine the optimal center for the large circle that maximizes the number of covered trees, as shown in Fig. \ref{SetCover}(a). The best circle that ultimately covers the most trees is added to the waypoint list 
$\textbf{\emph{W}}$. Then, the nearest uncovered tree is selected as the next starting point $\textbf{\emph{P}}_{nearest}$ as shown in Fig. \ref{SetCover}(b). The algorithm repeats this process until all small trees are fully covered. During the process, the potential points are within a search region centered at the $\textbf{\emph{P}}_{nearest}$, with side lengths defined by [$R$-$r$]. This greedy approach provides a fast solution, though it does not guarantee globally optimal results.
Finally, the generated waypoints would be used as input for the target-based path plan, as shown in Fig. \ref{SetCover}(d).

\begin{figure}[!t]
    \centering
    \includegraphics[width=4.5in]{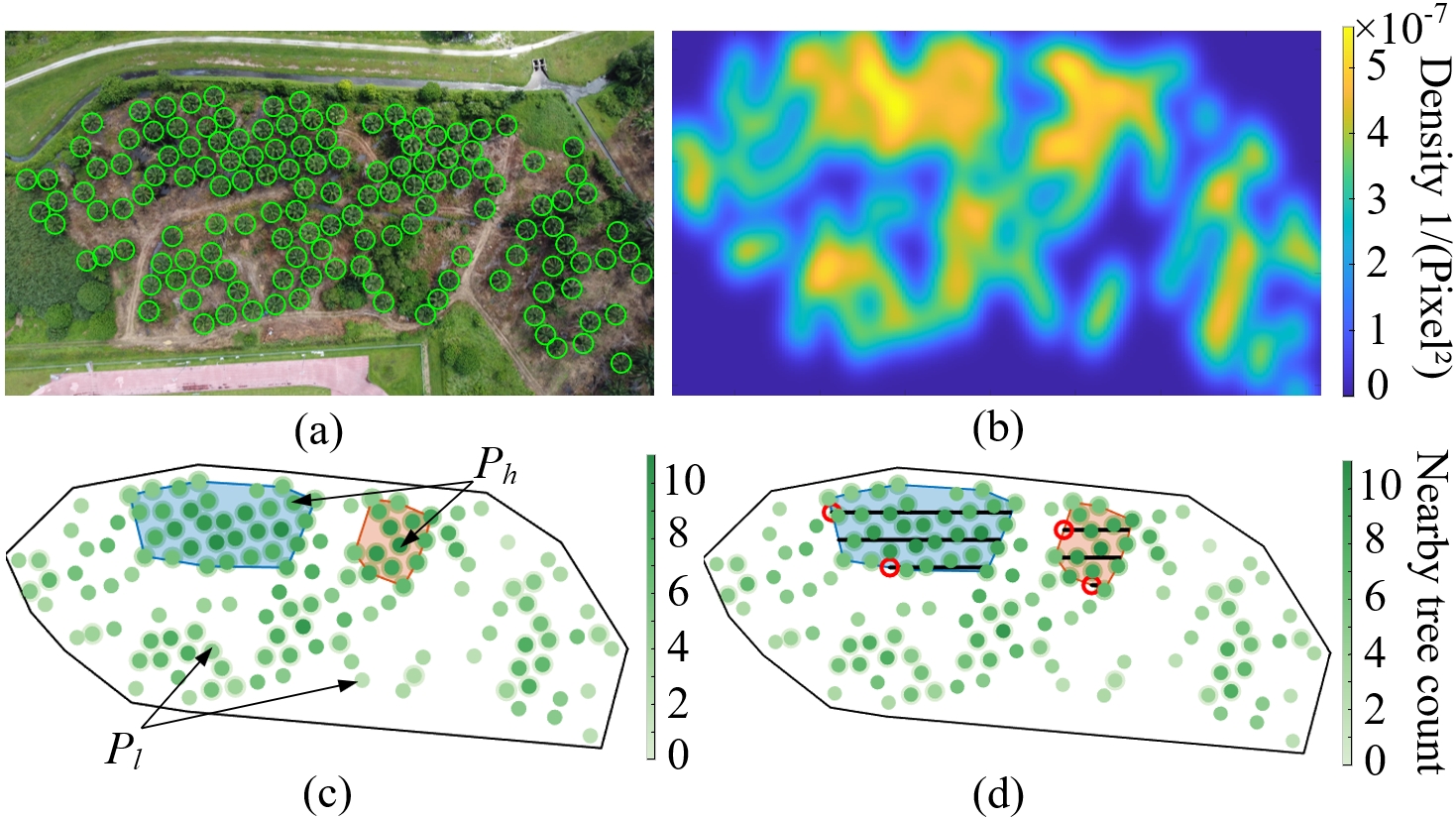}
    \caption{Dense region generation. (a) Tree detection results; (b) Kernel Density Estimation; (c) Dense region generation; (d) Boustrophedon path plan.}
    \label{DenseRegion}
\end{figure}

\begin{algorithm}[H]
\caption{Greedy Coverage Algorithm}
\begin{algorithmic}[1]
\STATE {\textbf{Input:}} Oil palm tree coordinates 
\STATE {\textbf{Output:}} Waypoints coordinates \(C\)
\STATE {Set the start point $\textbf{\emph{P}}_{start}$}
\STATE {\textbf{while not}} all small circles are covered \textbf{do}
\STATE \hspace{0.5cm} {\textbf{for}} each potential position around $\textbf{\emph{P}}_{start}$
\STATE \hspace{1cm} Add the $\textbf{\emph{P}}_{start}$ to waypoint list$\textbf{\emph{W}}$
\STATE \hspace{1cm} Calculate the number of covered trees
\STATE \hspace{1cm} {\textbf{if}} the large circle in this position covers more trees
\STATE \hspace{1.5cm} Replace the current point in $\textbf{\emph{W}}$
\STATE \hspace{1cm} {\textbf{end if}} 
\STATE \hspace{0.5cm} {\textbf{end for}}
\STATE \hspace{0.5cm} Find the nearest uncovered tree and set as $\textbf{\emph{P}}_{nearest}$
\STATE \hspace{0.5cm} Set $\textbf{\emph{P}}_{nearest}$ as the new $\textbf{\emph{P}}_{start}$
\STATE {\textbf{end while}}
\end{algorithmic}
\label{alg1}
\end{algorithm}
\newpage
However, this process is inefficient and unnecessary in regions with high tree density, as shown in Fig. \ref{DenseRegion}(a). We introduce a density-aware waypoint generation strategy that identifies dense regions using Kernel Density Estimation (KDE) to address this. It estimates the density distribution of trees as shown in Fig. \ref{DenseRegion}(b). This allows the classification of points into high-density \(P_h\) and low-density \(P_l\) as shown in Fig. \ref{DenseRegion}(c). 
DBSCAN algorithm is applied to \(P_h\) to segment the high-density regions further. DBSCAN clusters points by considering a point as a core point if at least a specified minimum number of neighbors exist within a given radius \(\epsilon\).
Finally, as shown in Fig. \ref{DenseRegion}(d), for each identified cluster \(R_k\), Boustrophedon Path Planning is then applied to generate a systematic back-and-forth sweeping path \(B_k\). This method simplifies path computation and enhances coverage consistency, making it effective in dense tree environments. The detail of the dense region generation is organized in Algorithm \ref{alg2}.

\begin{algorithm}[H]
\caption{Density-Based Region Extraction and Coverage Path Planning}
\begin{algorithmic}[1]
\STATE \textbf{Input:} Point set $P$, bandwidth $b$, clustering parameters $(\epsilon, \text{minPts})$, path width $w$
\STATE \textbf{Output:} Planned coverage paths over dense regions
\STATE Get density estimates $D$ using KDE with bandwidth $b$
\STATE Assign each point $p \in P$ a density value from $D$
\STATE Separate points into high-density $P_h$ and low-density $P_l$ 
\STATE Cluster $P_h$ using DBSCAN $(\text{minPts})$
\STATE Merge $P_h$ and $P_l$ separately

\FOR{each cluster $R_k$ in $P_h$}
    \STATE Compute convex hull $H_i$ of $R_k$
    \STATE Generate Boustrophedon paths $B_k$ for $H_i$ with width $w$
\ENDFOR
\end{algorithmic}
\label{alg2}
\end{algorithm}

\subsection{Target-based path planning}
\label{sec:PathPlan}
After the waypoints are generated in the previous step, the CPP problem can be transformed into the TSP. This transformation forms the essential foundation for implementing target-based path planning. In this study, turning angles and intersection counts were also considered in addition to travel distance. Minimizing travel distance and turning angles helps extend flight range and endurance within limited battery life. Reducing intersection counts improves efficiency, as seen in fruit harvesting, where fewer intersections result in saved resources \cite{hoffmann2024optimal}. We selected three methods based on common algorithm types to evaluate and demonstrate the generalization of the proposed CPP framework. These methods are Greedy Heuristic Insertion (GHI) \cite{voigt2024review} as a heuristic method, Ant Colony Optimization (ACO) \cite{li2023uav} as a swarm intelligence algorithm, and Monte Carlo Reinforcement Learning (MCRL) \cite{jevzek2024krrf, chaves2024parallel}.

\subsubsection{Greedy Heuristic Insertion}
As shown in Algorithm \ref{alg3}, this algorithm was chosen for its simplicity and efficiency. The core idea is to iteratively build the path by inserting the nearest points at each step \cite{voigt2024review}. A random starting point is selected, and the two closest points to the starting point are identified to form the initial path. Subsequently, the remaining points are inserted into the path based on the established waypoints. For each insertion, the algorithm calculates the changes in distance, angle, and intersection counts caused by placing the point between adjacent nodes. The combined changes, as shown in Eq. (\ref{EQ1}), serve as the objective function $\emph{L}$ guiding the point insertion process.

\begin{equation}
\emph{L} = \alpha_{GHI} \cdot d + \beta_{GHI}  \cdot \delta + \omega_{GHI}  \cdot n,
\label{EQ1}
\end{equation}
where $\emph{d}$ is the total travel distance, 
$\delta$ represents the cumulative turning angles in the path, and 
$\emph{n}$ accounts for intersection counts. The parameters $\alpha_{GHI}$, $\beta_{GHI}$, and $\omega_{GHI}$ are weighting factors that balance the importance of path length, turning angle, and intersection counts in the final cost evaluation.
To ensure the optimization of the path, the algorithm evaluates each possible insertion of a new point by computing combined score changes before and after the insertion. This process continues until all points are inserted into the path.

\begin{algorithm}[H]
\caption{Greedy Heuristic Insertion}
\begin{algorithmic}[1]
\STATE {\textbf{Input:}} Waypoints coordinates \(C\)
\STATE {\textbf{Output:}}  Optimized path \(P^*\)

\STATE Initialize path \( P \) with a random starting waypoint \( s_0 \)
\STATE Find the two nearest waypoints \( n_1, n_2 \) to \( s_0 \) and add to \( P \)

\WHILE{Remaining waypoints exist}
\STATE Randomly select a new waypoint \( p_{\text{new}} \)
\STATE Insert \( p_{\text{new}} \) into \( P \) at the position minimizing the combined distance \(d\), angle \(\delta\) and intersection counts \(n\).
\ENDWHILE

\STATE Return the optimized path \( P^* = P \)
\end{algorithmic}
\label{alg3}
\end{algorithm}

\subsubsection{Ant Colony Optimization}
In solving the TSP, ACO can explore large solution spaces efficiently \cite{li2023uav}. Inspired by the behavior of ants, ACO effectively balances exploration and exploitation, enabling it to consider various constraints, such as travel distance and turning angles. 
In this study, there are 
$\emph{n}$ waypoints. Each ant starts from the first randomly selected waypoint and selects the next waypoint based on the pheromone concentration along the path, continuing until all waypoints have been visited. After each iteration, the model records the path cost for each ant and outputs the path with minimal cost in all iterations.
As shown in Algorithm \ref{alg4}, it proceeds as follows:

Initialization: The algorithm takes as input a set of waypoints \(C\), along with parameters: the pheromone importance \(\alpha_{ACO}\), heuristic importance \(\beta_{ACO}\), angle weight \(\gamma_{ACO}\), intersection weight \(\omega_{ACO}\), pheromone decay rate \(\rho\), pheromone constant \(Q\), number of ants \(m\), and iterations \(N\). The distance matrix \(D\) is normalized. The pheromone matrix \(\tau(i,j)\) is initialized uniformly, where \(i\) and \(j\) represent indices of waypoint pairs in \(C\).

Path Construction: Each ant begins from a randomly selected waypoint \(s_0\), forming an initial path \(P\) with a visited set \(V_d\). The ant iteratively selects the next waypoint based on a probability distribution defined as:
\begin{equation}
p(i, j) = \frac{A(i, j)}{\sum_{k \in \text{unvisited}} A(i, k)},
\end{equation}
where \(A(i, j)\) represents the attractiveness of node \(j\), combining pheromone intensity, heuristic distance information, and geometric constraints:
\begin{equation}
A(i, j) = \tau(i, j)^{\zeta_{ACO}} \eta(i, j)^{\alpha_{ACO}} \delta_{ACO}(i, j)^{\beta_{ACO}} n_{ACO}(i, j)^{\omega_{ACO}},
\end{equation}
where \(\eta(i, j)\) represents heuristic information, defined as the inverse distance between waypoints; \(\delta_{ACO}(i, j)\) is the turning angle reward calculated from the turning angle formed by the previous point, current point, and candidate point \(j\), penalizing sharp turns; \(n_{ACO}(i, j)\) is the intersection reward which counts the number of intersections that would be introduced by adding the waypoint \((i, j)\)  to the current path; \(\alpha_{ACO}, \beta_{ACO}, \gamma_{ACO}, \zeta_{ACO}\) are tunable parameters that control the relative importance of each factor.
Waypoint selection follows an \(\epsilon\)-greedy policy. With the exploration rate \(\epsilon\), a random unvisited waypoint is selected; otherwise, the waypoint with highest probability is chosen:
\begin{equation}
j = \arg\max_{j'} p(i, j').
\label{eq_greedy}
\end{equation}
Here, \(i\) is the current waypoint, \(j'\) denotes all feasible next waypoints, and \(j\) is the selected waypoint with the highest transition probability \(p(i, j')\).
The selected waypoint \(j\) is added to \(P\) and the visited set \(V_d\). This process continues until all waypoints are visited.

Pheromone Update: After generating a complete path, the algorithm evaluates its total cost, incorporating distance, Angle reward, and Intersection reward. Based on this evaluation, the pheromone matrix is then updated, starting with pheromone evaporation:
\begin{equation}
\tau(i, j) = (1 - \rho) \tau(i, j),
\end{equation}
where \(\rho\) is the pheromone evaporation rate.
Then, pheromones are deposited based on the path quality:
\begin{equation}
\tau(i, j) = \tau(i, j) + \sum_{m} \frac{Q}{L_m},
\end{equation}
where \(L_m\) is the total length of the path found by ant \(m\), \(Q\) is the total amount of pheromone. To encourage more exploration in early iterations and greater exploitation of learned pheromone information later, the algorithm adjusts the exploration rate \(\epsilon\) over iteration \(i\):
\begin{equation}
\epsilon = \max(\epsilon_{\min}, \epsilon_{\max} \cdot e^{-\lambda i}).
\label{eq:epsilon_decay}
\end{equation}

\subsubsection{ Monte Carlo Reinforcement Learning}
As shown in Algorithm \ref{alg5}, this hybrid algorithm combines the Monte Carlo method \cite{jevzek2024krrf} with the Q-table for path optimization. The Monte Carlo method enables global trajectory evaluation by sampling complete paths, thereby accounting for the overall impact of previous and future states. Meanwhile, the Q-table originally derived from state-action function in Q-learning serves here as a lookup table to store and retrieve the expected utility of state–action pair. In this context, the state space \(S\) corresponds to the set of waypoints that have already been included in the current path. Each state represents a partial sequence of visited waypoints, reflecting the current progress of the path construction. The action space 
\(a\)  corresponds to the set of waypoints that have not yet been included, representing the candidate waypoints for the next step. Specifically, the current state \(s\) aligns with the previously defined set 
\(i\), denoting the indices of waypoints already selected, while each action 
\(a\)  corresponds to an element in the set \(j\) , which represents the indices of unvisited waypoints. The algorithm proceeds as follows:

Initialization: A set of waypoints \(C\) is provided as input, along with the discount factor \(\gamma_{MC}\), explore rate \(\epsilon\), learn rate \(\eta_{MC}\), travel distance weight \(\alpha_{MC}\), angle weight \(\beta_{MC}\), intersection weight \(\omega_{MC}\). The Q-table \(Q\) and reward matrix \(R\) are also initialized. The distance matrix \(D\) and reward matrix \(R\) are computed based on the pairwise distances between waypoints.

Path generation and optimization: For each episode, the algorithm randomly selects an initial waypoint \(s_0\) and initializes the visited set \(V_d\) and the path \(P\). The algorithm then iteratively selects the next waypoint \(a\) using an \(\epsilon\)-greedy policy as the same in Eq. (\ref{eq_greedy}). The selected waypoint \(a\) is added to visited and appended to \(P\), and the current state \(s\) is updated to \(a\). This process continues until all waypoints are visited. Finally, the path is closed by appending the initial waypoint \(s_0\) to \(P\), forming a closed loop.

\begin{algorithm}[H]
\caption{Ant Colony Optimization}
\begin{algorithmic}[1]
\STATE \textbf{Input:} Waypoint coordinates \(C\)

\STATE \textbf{Output:} Optimized path \(P^*\)

\STATE Compute distance matrix \(D\) and normalize it
\STATE Initialize pheromone matrix \(\tau(i, j) = 1\) for all edges

\FOR{iteration \(i = 1\) to \(N\)}
    \FOR{each ant \(m = 1\) to \(M\)}
        \STATE Select random starting waypoint \(s_0\), initialize path \(P = [s_0]\), visited set \(V_d = \{s_0\}\)
        
        \WHILE{\(|V_d| < n\)}
            \STATE Compute transition probabilities \(p(i, j)\) using pheromone and heuristic information
            \STATE Select next waypoint \(a\) using \(\epsilon\)-greedy strategy:
            \STATE \hspace{0.5cm} With probability \(\epsilon\), select random unvisited waypoint
            \STATE \hspace{0.5cm} Else, select \(j = \arg\max_{j'} p(i, j')\)
            \STATE Append \(j\) to \(P\) and update \(V_d\)
        \ENDWHILE
        \STATE Close path: Append \(s_0\) to \(P\)
    \ENDFOR
    
    \FOR{each ant \(m\)}
        \STATE Total distance \(d_{ACO}\) of path \(P_m\)
        \STATE Angle reward \(\delta_{ACO} = 1/(1 + \delta)\)
        \STATE Intersection reward \(n_{ACO} = \exp(-\mu \cdot n)\)
        \STATE \(R_m = \alpha_{quality} \cdot d_{ACO} + \beta_{quality} \cdot \delta_{ACO} + \omega_{quality}  \cdot n_{ACO}\)
        
        \IF{\(R_m\) < \(R^*\)}
        \STATE \(P^* = P_m\)
        \ENDIF
    \ENDFOR
    
    \STATE Pheromone evaporation and deposition.
    \STATE Update \(\epsilon\)
\ENDFOR

\STATE Return the optimized path \( P^* \)
\end{algorithmic}
\label{alg4}
\end{algorithm}

Cumulative reward calculation: After generating a complete path, the algorithm calculates the cumulative reward. The cumulative reward \(G_t\) and intersection count \(n\) are initialized to zero. For each step \(t\) in the path (starting from the last waypoint and moving backward), the immediate reward \(r_t\) is computed:
\begin{equation}
 r_t = \alpha_{MC} \cdot d_{MC} + \beta_{MC} \cdot \delta_{MC} + \omega_{MC} \cdot n_{MC},
\end{equation}
where \(\alpha_{MC}, \beta_{MC}\) and \(\omega_{MC}\) are weights for the immediate distance reward \(d_{MC}\), angle reward \(\delta_{MC}\), and intersection reward \(n_{MC}\), respectively. And \(\gamma_{MC}\) is the discount factor for future rewards. The angle reward \(\delta_{MC}\) is calculated based on the turning angle, and the intersection reward \(n_{MC}\) is computed based on the intersections counts \(n\). 
The cumulative discounted reward \(G_t\) is then computed recursively as
\begin{equation}
 G_t = r_t + \gamma_{MC} \cdot G_{t+1}.
\end{equation}
Finally, the Q-value for the current state-action pair is updated using
\begin{equation}
Q(s_t, a_t) = Q(s_t, a_t) + \eta_{MC} \cdot (G - Q(s_t, a_t)).
\end{equation}
The parameter \(\eta_{MC}\) is the learning rate, controlling how much the sampled return \(G\) influences the Q-value update:
\begin{equation}
\eta_{MC} = \frac{1}{visitCounts(s, t)},
\end{equation}
where \(visitCounts(s, t)\) tracks how many times a particular state-action pair has been visited. The learning rate \(\eta_{MC}\) decreases as the visit count increases, which ensures that Q-value updates become smaller and more stable as more experiences are accumulated.
Meanwhile, the exploration rate \(\epsilon\) decays over time using an exponential decay formula, as is the same as in Eq. (\ref{eq:epsilon_decay}). This ensures that the algorithm explores more in the early stages and exploits learned knowledge in the later stages.

\begin{algorithm}[H]
\caption{Monte Carlo Reinforcement Learning}
\begin{algorithmic}[1]
\STATE \textbf{Input:} Waypoint coordinates \(C\)

\STATE \textbf{Output:} Optimized path \(P^*\)

\STATE Init \(Q(s, a) = 0\)

\STATE Compute distance matrix \(D\)

\FOR{episode \(= 1\) to \(N\)}
\STATE Initial state \(s_0\), set visited set \( V_d = \{s_0\}\), path \(P = [s_0]\)

\WHILE{\(|V_d| < n\)}
\STATE Pick action \(a\) using \(\epsilon\)-greedy:
\STATE \hspace{0.5cm} With prob \(\epsilon\), random unvisited \(a\)
\STATE \hspace{0.5cm} Else, \(a = \arg\max_{a'} Q(s, a')\)
\STATE Add \(a\) to \(\text{visited}\), append to \(P\), update \(s \leftarrow a\)
\ENDWHILE
\STATE Close path: Append \(s_0\) to \(P\)
\STATE Init cumulative reward \(G_{t+1} = 0\)
\FOR{\(t = |P|\) to \(1\)}
\STATE Reward \(r_t = \alpha_{MC} \cdot d_{MC} + \beta_{MC} \cdot \delta_{MC} + \omega_{MC} \cdot n_{MC}\)
\STATE Angle reward \(\delta_{MC} = (1 - \delta / 180)^2\)
\STATE Intersection reward \(n_{MC}= \exp(-\zeta_{MC} \cdot n)\)
\STATE Update \( G_t = r_t + \gamma_{MC} \cdot G_{t+1}.\)
\STATE Update \(Q(s_t, a_t)\)
\ENDFOR
\STATE Update \(\epsilon\)
\STATE \(R = \alpha_{quality} \cdot d_{MC} + \beta_{quality}  \cdot \delta_{MC} + \omega_{quality}  \cdot n_{MC}\)
\IF{\(R\) < \(R^*\)}
\STATE \(P^* = P\)
\ENDIF
\ENDFOR

\STATE Return the optimized path \( P^*\)
\end{algorithmic}
\label{alg5}
\end{algorithm}

\subsection{Object-optimized path replanning}
\label{sec:RePlan}
Although the waypoints have been set, they are not the only positions within the coverage circle that can cover the trees inside. Each waypoint defines a feasible region where alternative positions can also effectively cover the same trees, offering flexibility in path planning. As long as the MAV is within this region, all trees within the coverage area can still be observed. To further optimize the planned path, we propose an object-optimized replanning approach that refines the path planning by selecting the best points within the feasible region rather than strictly following the set waypoints.

\begin{figure}[!t]
    \centering
    \includegraphics[width=4.5in]{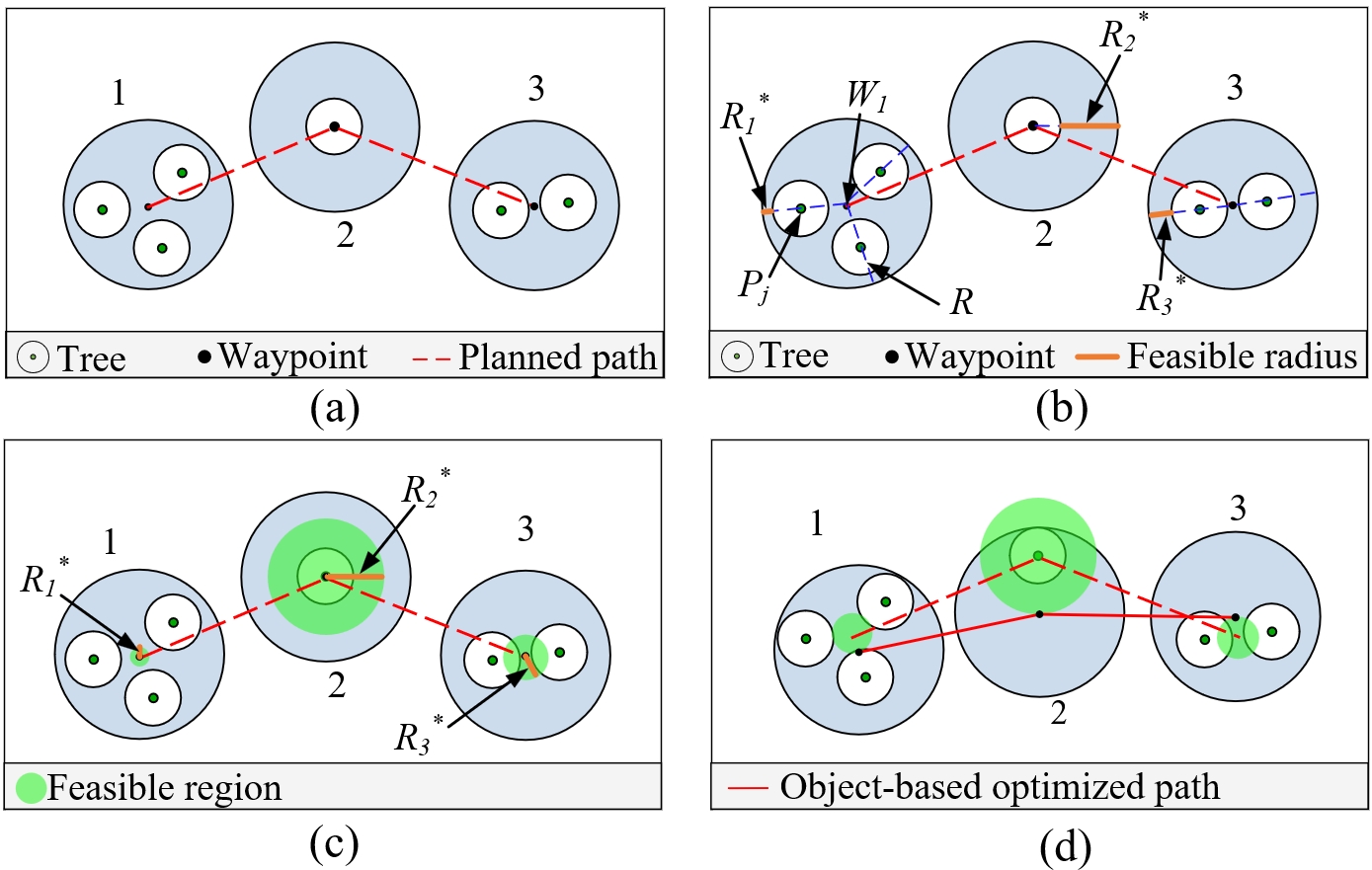}
    \caption{Object-optimized path replanning. (a) Initial waypoints and planned path; (b) Feasible radius calculate; (c) Feasible region generate; (d) Path replan.}
    \label{PathReplan}
\end{figure}

As shown in Fig. \ref{PathReplan} and Algorithm \ref{alg6}, the key idea is to optimize the path by considering both the distance and angle between successive points. The path can be further optimized by selecting the optimal waypoint within the feasible region, resulting in a more efficient CPP strategy. The algorithm proceeds as follows:

Initialization: The path \(P^*\) is initialized with waypoint coordinates \(C\), where each point \(W_i\) on the original path corresponds to a waypoint as shown in Fig. \ref{PathReplan} (a).

Feasible region generates: As shown in Fig. \ref{PathReplan} (b) and (c), The feasible region of each \(W_i\) is defined by the feasible radius \(R^*_i\), which could be calculated by:
\begin{equation}
R^*_i = R - \max_{j \in S_i} \| P_j - W_i \| - r ,
\end{equation}
where \(R\) is the radius of the coverage circle, \(P_j\) is the coordinates of trees \(S_i\) which belong to the coverage circle of the \(W_i\), and \(r\) is the tree crown radius.

Object-Optimized Replan: For each waypoint \(W_i\), the algorithm identifies optimal points within the feasible region defined by its coverage circle. Specifically, It evaluates candidate points by computing the total travel distance and turning angle based on \((W_{i-1}, W_i, W_{i+1})\).

Path Update: As shown in Fig. \ref{PathReplan} (d), the path \(P_{optimal}\) is updated with the newly selected waypoints. This process continues until all waypoints are refined. This object-optimized replanning approach enhances MAV navigation efficiency by reducing unnecessary travel while maintaining tree detection.

\begin{algorithm}[H]
\caption{Object-Optimized Path Replanning}
\begin{algorithmic}[1]
\STATE \textbf{Input:} Original planned path \( P^* \), waypoint coordinates \( C \)
\STATE \textbf{Output:} Optimized path \( P_{optimal} \)

\FOR{each waypoint \( W_i \) in \( P^* \)}
    \STATE Generate the feasible region for \( W_i \) 
    \STATE Evaluate points within the feasible radius \(R^*_i\) by comparing distance and angle relative to \( W_{i-1} \) and \( W_{i+1} \)
    \STATE Select the point that minimizes the distance and angle
    \STATE Update \( P_{optimal} \) with the selected point
\ENDFOR

\STATE Return the optimized path \( P_{optimal} \)
\end{algorithmic}
\label{alg6}
\end{algorithm}

\begin{table}
\begin{center}
\renewcommand{\arraystretch}{1.5} 
\caption{Main parameters used in algorithms.}
\label{tab1}
\begin{tabular}{ m{3cm}  m{8.5cm}  m{2cm}}
\toprule

\multirow{2}{*}{Parameters} & \multicolumn{2}{c}{\textbf{Global}} \\
\cline{2-3}
 & Explanation & Value \\
\midrule
\( R \) & The field of view of the camera & 175  \\
\( r \) & The radius of the oil palm tree & 50  \\
\( b \) & The bandwidth in  KDE & 80  \\
\(\alpha_{quality}\) & Travel distance weight in the evaluation of the path quality & 0.3 \\
\(\beta_{quality}\) & Turning angle weight in the evaluation of the path quality & 0.7  \\
\(\omega_{quality}\) & Intersection weight in the evaluation of the path quality & 500  \\
\midrule

\multirow{2}{*}{Parameters} & \multicolumn{2}{c}{\textbf{Greedy Heuristic Insertion}} \\
\cline{2-3}
 & Explanation & Value \\
\midrule
\(\alpha_{GHI}\) & Travel distance weight in GHI & 0.3  \\
\(\beta_{GHI}\) & Turning angle weight in GHI & 0.7  \\
\(\omega_{GHI}\) & Intersections weight in GHI & 500  \\
\midrule

\multirow{2}{*}{Parameters} & \multicolumn{2}{c}{\textbf{Ant Colony Optimization}} \\
\cline{2-3}
 & Explanation & Value \\
\midrule
\(\alpha_{ACO}\) & Heuristic information (travel distance) weight in ACO & 1  \\
\(\beta_{ACO}\) & Turning angle weight in ACO & 1  \\
\(\omega_{ACO}\) & Intersections weight in ACO & 4  \\
\(\zeta_{ACO}\) & Pheromone weight in ACO & 1 \\
\(\rho\) & The evaporation rate of pheromone in ACO & 0.3 \\
\(Q\) & The total amount of pheromone in ACO & 1000 \\
\(m\) & The amount of ants in ACO & 100 \\
\midrule

\multirow{2}{*}{Parameters} & \multicolumn{2}{c}{\textbf{Monte Carlo Reinforcement Learning}} \\
\cline{2-3}
 & Explanation & Value \\
\midrule
\(\alpha_{MC}\) & Travel distance weight in MC & 0.3  \\
\(\beta_{MC}\) & Turning angle weight in MC & 0.7  \\
\(\omega_{MC}\) & Intersection weight in MC & 500  \\
\(\gamma_{MC}\) & The learn rate in MC & 0.2  \\
\midrule

\end{tabular}
\end{center}
\end{table}

\section{Experiments and discussion}
\label{sec:results}
This section evaluates the proposed CPP framework in a cluttered plantation with dispersed and irregularly distributed plants. Three algorithms are used to evaluate our framework. To further validate the framework in the real-world application, we set up a plantation scenario and conducted flight experiments. The results of multiple optimization steps applying the three algorithms are presented, followed by a performance comparison. In addition, the findings of flight experiments are demonstrated and analyzed. The value of the main parameters used in these algorithms is shown in Table \ref{tab1}.






















\subsection{Performance evaluation of optimization steps}
The selected cluttered plantation site features irregularly distributed palm trees, as illustrated in Fig. \ref{Flowchart}. In addition to palm trees, the area also contains non-target vegetation, further increasing the complexity of the environment. The density of palm tree distribution varies across the site, forming distinct high-density and low-density zones. To extract tree coordinates within the plantation, a HOG-based palm tree detection method is first applied, as shown in Fig. \ref{DenseRegion}(a). Next, Algorithm \ref{alg2} is used to identify high-density plantation regions, where Boustrophedon path plan is employed to generate a coverage path. For the remaining trees outside these high-density regions, Algorithm \ref{alg1} produces multiple coverage circles to determine waypoint coordinates, transforming the CPP problem into a TSP problem.

\begin{figure}[!t]
    \centering
    \includegraphics[width=4.5in]{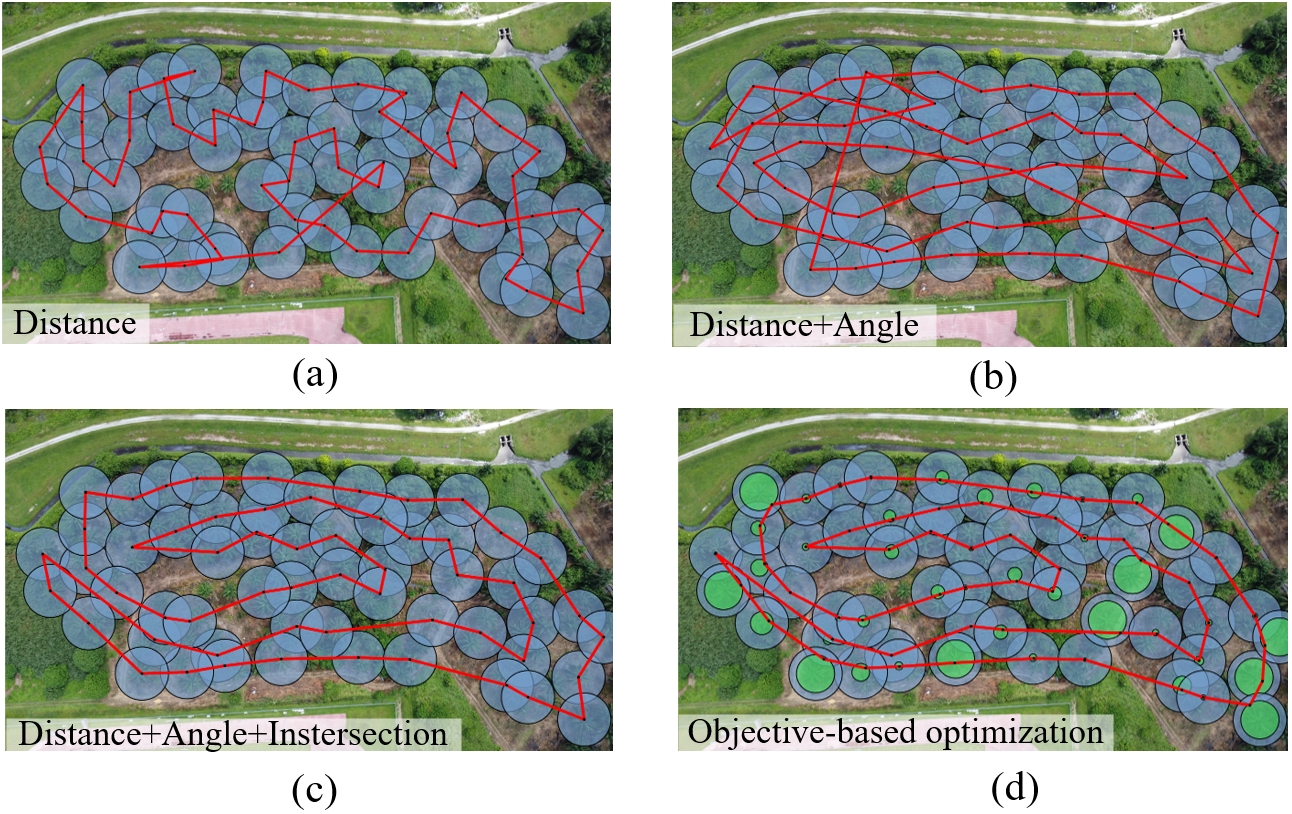}
    \caption{Path planning results using the MCRL Method in different optimization steps: (a) Path optimized for distance; (b) Path optimized for both distance and turning angle; (c) Path optimized for distance, angle, and intersection count; (d) Object-based optimization.}
    \label{MCRL}
\end{figure}

To clearly demonstrate the performance improvements at each optimization step, the path plan results using the MCRL method are analyzed first.
Fig. \ref{MCRL} (a) represents the initial path optimized solely for distance, resulting in the short path length. However, this results in sharp turns, with a high cumulative turning angle of 4042 degrees and four path intersections. Fig. \ref{MCRL} (b) incorporates turning angle optimization, which significantly smooths the path, reducing the cumulative turning angle by approximately 50 \%. However, this comes at the cost of a 39 \% increase in total path length and an increase in path intersections to 10. Fig. \ref{MCRL} (c) further optimizes the path by considering path intersections, eliminating all intersections while keeping the total path length increase minimal (about 10 \% compared to (a)). Meanwhile, the cumulative turning angle is further reduced to 2016 degrees. Fig. \ref{MCRL} (d) applies object-based optimization, as shown in Algorithm \ref{alg6}. This results in a 16.9 \% reduction in path length compared to Fig. \ref{MCRL} (c) and a further decrease in the angle to 1650 degrees. The final result shows that optimizing for all objectives yields a more efficient and smoother path.

\subsection{Comparison of Algorithm Performance}
\begin{figure}[H]
    \centering
    \includegraphics[width=4.5in]{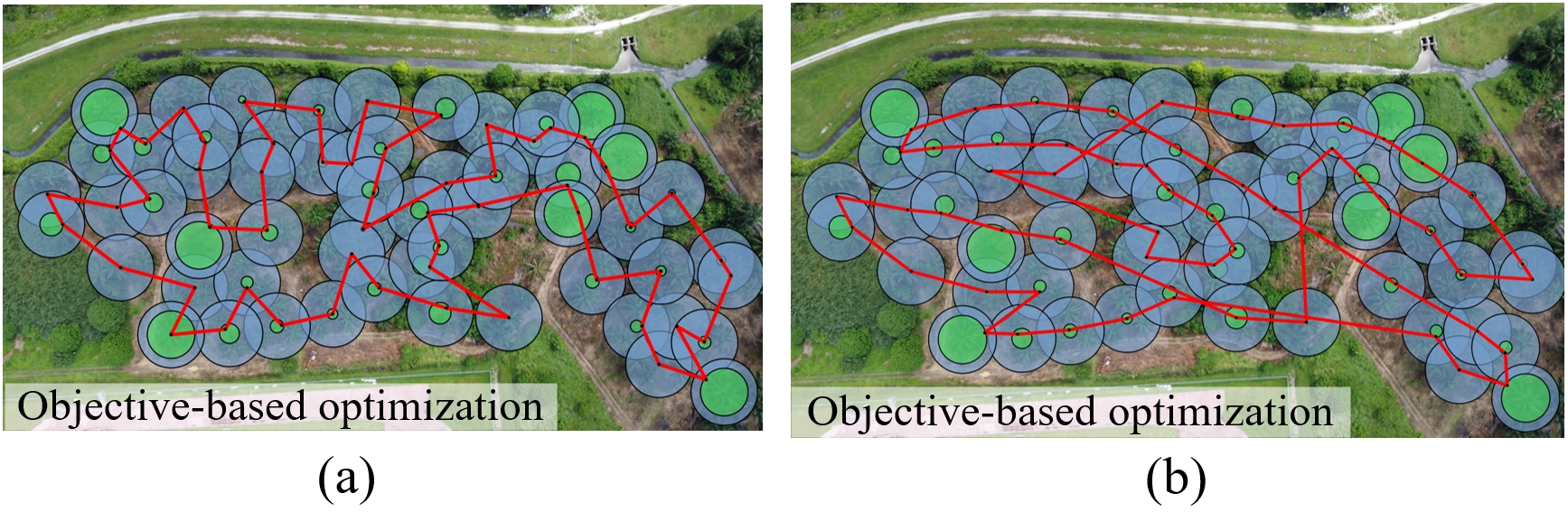}
    \caption{Path plan results by (a) GHI and (b) ACO algorithms under object-based optimization.}
    \label{ACO_GHI}
\end{figure}

\begin{figure}[H]
    \centering
    \includegraphics[width=4.5in]{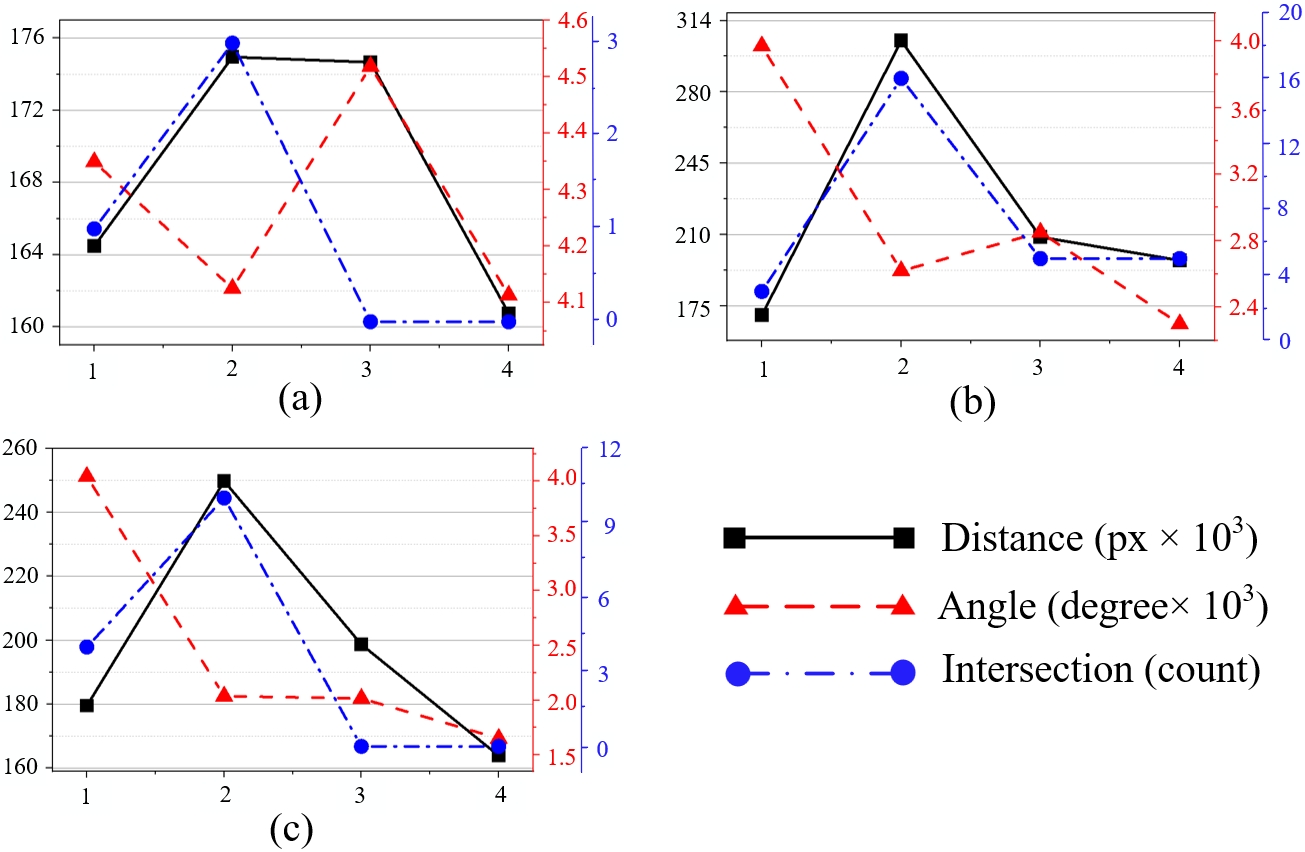}
    \caption{Path plan results using different algorithms: (a) GHI; (b) ACO; (c) MCRL. The X-axis represents different optimization goals: (1) Distance; (2) Distance and angle; (3) Distance, angle, and intersection count; (4) Object-based optimization.}
    \label{DataResult}
\end{figure}

To validate the generalizability of the proposed framework, it has also been applied to the other two algorithms, GHI and ACO, as shown in Fig. \ref{ACO_GHI}. The path planning results using different algorithms—GHI, ACO, and MCRL—are compared in Fig. \ref{DataResult}. The evaluation focuses on total path length, cumulative turning angle, and the number of path intersections.

\begin{itemize}
\item GHI Algorithm: The GHI algorithm achieves relatively short path lengths across all cases. However, it results in higher turning angles, with values exceeding 4000 degrees in most cases. Meanwhile GHI maintains minimal path intersections, ranging from 0 to 3. The object-based optimization of the flight path is shown in Fig. \ref{ACO_GHI} (a).

\item ACO Algorithm: The ACO algorithm demonstrates a significant trade-off between path length and smoothness. While it achieves a lower turning angle (minimum 2300 degrees), it often results in increased path lengths, with some cases showing over an 80 \% increase compared to GHI. The object-based optimization of the flight path is shown in Fig. \ref{ACO_GHI} (b). Additionally, ACO causes more path intersections, reaching a maximum of 16, which may lead to inefficiencies in real-world applications.

\item MCRL Algorithm: The MCRL method effectively balances all objectives. While its initial path length is slightly longer than GHI and ACO, it significantly reduces the turning angle, reaching as low as 1650 degrees after optimization. Moreover, MCRL eliminates path intersections in its final results, demonstrating superior efficiency in generating smooth and feasible paths.
\end{itemize}

Overall, combined with the numerical results in Table \ref{tab2}, the comparison shows that GHI provides the shortest paths. However, it still suffers from excessive turning angles despite considering angle minimization. This is because its greedy heuristic insertion strategy optimizes distance, angle, and intersections locally at each insertion step rather than globally over the entire path. As a result, while individual waypoint insertions aim to reduce sharp turns, the lack of a global smoothing mechanism leads to suboptimal transitions between waypoints, resulting in high cumulative turning angles. ACO enhances smoothness by utilizing pheromone-based learning, enabling the algorithm to explore multiple potential paths before converging on an optimized solution.

\begin{table}[H]
\begin{center}
\renewcommand{\arraystretch}{1.5} 
\caption{Comparison of path plan results based on object-based optimization using three algorithms.}
\label{tab2}
\begin{tabular}{ m{1.5cm}  c  c  c }
\toprule
Methods & Distance & Angle & Intersection \\
\midrule
GHI & 16075 & 4113 & 0 \\

ACO & 19733 & 2300 & 5 \\

MCRL & 16386 & 1650 & 0 \\
\bottomrule
\end{tabular}
\end{center}
\end{table}

 However, its heuristic-based waypoint selection is influenced by accumulated pheromone values, which are updated based on overall path quality rather than immediate local constraints. This can lead to longer paths and increased intersections, as the algorithm balances exploration and exploitation without directly optimizing for intersection minimization. MCRL, on the other hand, offers a well-balanced solution by iteratively refining path selection through reinforcement learning. It provided a 16.9 \% shorter path than ACO while maintaining a path length comparable to GHI (1.9 \% longer than GHI). It also significantly improves trajectory smoothness, reducing the turning angle by 28.3 \% compared to ACO  and 59.9 \% compared to GHI. 

Additionally, MCRL eliminates intersections, matching GHI but outperforming ACO, which introduces five intersections. Instead of relying on indirect pheromone reinforcement, MCRL directly incorporates penalties for sharp turns and intersections into its reward function, allowing it to gradually learn a more optimal balance between path length, smooth turns, and minimal intersections. By continuously updating its policy based on cumulative rewards, MCRL achieves a globally optimized trajectory, making it a more compelling choice for path planning.

\subsection{Flight experiments}
\begin{figure}[!t]
    \centering
    \includegraphics[width=3in]{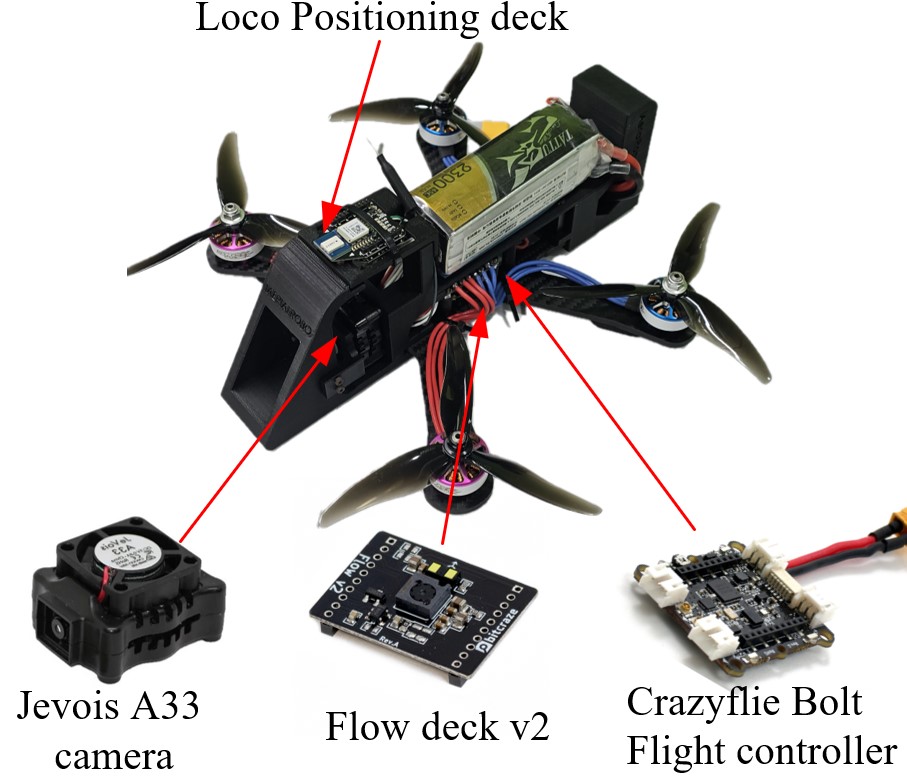}
    \caption{The prototype of the customized MAV.}
    \label{mav}
\end{figure}

To validate the proposed framework, we constructed a cluttered plantation environment inspired by real-world conditions, reflecting key characteristics of dispersed and irregular plantations. As shown in Fig. \ref{mav}, A custom-built MAV was used for the experiments, equipped with a Crazyflie Bolt 1.1 flight controller, a Flowdeck V2 deck, a Loco Positioning deck, and a JeVois-A33 camera. The camera detects oil palm trees onboard and transmits their image positions to the flight controller for real-time navigation adjustments. During the flight, we implemented our previously developed ForaNav method, ensuring that the MAV followed the planned path while accurately positioning itself directly above each tree.
Separately, to plan the overall path and waypoints, we applied the proposed method offline on a computer, using tree detection results from the entire plantation map. This two-stage process allowed the MAV to follow a precomputed path and adapt its trajectory accurately during navigation to position itself above each detected tree.

\begin{figure}[!t]
    \centering
    \includegraphics[width=4.5in]{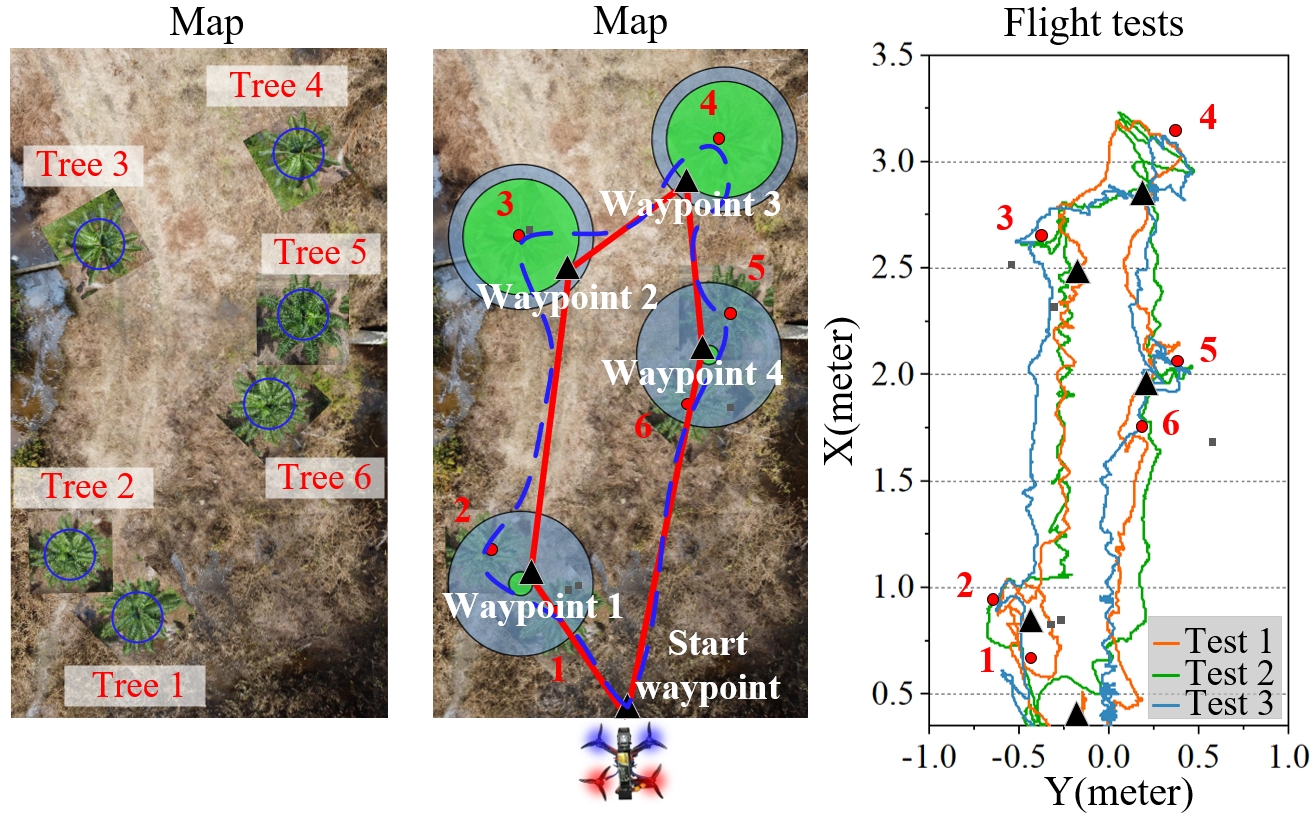}
    \caption{Flight experiment results in a cluttered environment. The left figure shows the plantation map with tree detection results. The middle figure illustrates waypoints and planned paths, where the blue dashed line represents the actual flight trajectory after applying ForaNav. The right figure shows the MAV trajectories from three flight tests.}
    \label{FlightPath}
\end{figure}

As shown in Fig. \ref{FlightPath}, the environment contains six trees, with Trees 1 and 2, as well as Trees 5 and 6, clustered together. Given the relatively sparse distribution of trees, we applied the GHI algorithm for path planning to generate a minimal and efficient route. As shown in the middle of Fig. \ref{FlightPath}, the resulting path consists of five waypoints, including one designated for takeoff and landing. Notably, Trees 1 and 2 were successfully assigned to the coverage area of Waypoint 1, while Trees 5 and 6 were covered by Waypoint 4. The red line represents the planned path under object-based optimization, while the blue line shows the actual flight path using ForaNav. The right side of Fig. \ref{FlightPath} illustrates the flight trajectory following the planned waypoints. By applying ForaNav, the MAV successfully visited all assigned waypoints while positioning itself precisely above each tree. This demonstrates the effectiveness of our method in ensuring both complete coverage and accurate positioning in real-world cluttered environments.

\section{Conclusions}
\label{sec:conclusion}
In this study, we proposed a multi-objective coverage path planning framework for addressing challenges in dispersed and irregularly planted regions. The framework effectively planned paths by integrating distance, turning angle, and intersection count. Additionally, a density-aware coverage method utilizing KDE and DBSCAN clustering was introduced to reduce the number of unnecessary waypoints. Furthermore, we developed an object-optimized path replanning strategy, allowing adaptive waypoint selection within feasible regions to enhance path efficiency. The integration of ForaNav enabled real-time tree detection and precise MAV positioning, facilitating accurate waypoint navigation for targeted interventions. Simulations and real-world flight experiments demonstrated the effectiveness of our proposed framework. Comparative analysis of different algorithms highlighted the advantages of the MCRL approach in achieving well-balanced path efficiency, minimizing total travel distance, reducing turning angles, and eliminating intersections. The flight experiments further validated the framework applicability in cluttered plantation environments, ensuring precise MAV positioning over target trees. The proposed framework provides a solution for optimizing MAV-based agricultural operations and planning the path for autonomous plantation tasks such as precision spraying and target monitoring. The project code and introduction are available in our repository\footnotemark[1].

\setcounter{footnote}{1}
\footnotetext{Repository:\url{ https://github.com/iAerialRobo/CPP-DIP.git}}

The current framework also opens opportunities for future enhancement by integrating multiple agents, potentially increasing coverage efficiency and scalability in large-scale plantations. The future extensions will also focus on equipping MAVs with onboard real-time coverage path planning capabilities. This advancement will enable greater adaptability to dynamic environments.

\section*{Acknowledgement} 
The authors would like to thank Malaysian Ministry of Higher Education (MOHE) for providing the Fundamental Research Grant Scheme (FRGS) (Grant number: FRGS/1/2024/TK04/USM/02/3).
The first author would like to thank the Chinese Scholarship Council (CSC) for the Ph.D. financial support (File No. 202106830031). 
\bibliographystyle{ieeetr}
\bibliography{Reference}

\end{singlespacing}

\end{document}